\documentclass[10pt,twocolumn,letterpaper]{article}

\usepackage{cvpr}
\usepackage{times}
\usepackage{epsfig}
\usepackage{graphicx}
\usepackage{amsmath}
\usepackage{amssymb}
\usepackage{siunitx}
\usepackage{subfigure}
\usepackage{multirow}
\usepackage{caption}
\graphicspath{{./images/}}

\def\D{\Delta}
\def\Re{\mathbb R}
\def\p{\varphi}
\def\1{\mathds{1}}

\def\ie{{i.e.$\;$}}

\newcommand{\X}{\mathcal{X}}
\newcommand{\Y}{\mathcal{Y}}

\newcommand{\myparagraph}[1]{\vspace{5pt}\noindent{\bf #1}}

\usepackage{microtype}

\usepackage[breaklinks=true,bookmarks=false]{hyperref}

\cvprfinalcopy % *** Uncomment this line for the final submission

\def\cvprPaperID{1855} % *** Enter the CVPR Paper ID here

\ifcvprfinal\fi
\begin{document}

%%%%%%%%% TITLE
\title{Gaze Embeddings for Zero-Shot Image Classification}

\author{
Nour Karessli$^1\thanks{Nour Karessli is currently with Eyeem, Berlin. The majority of this work was done in Max Planck Institute for Informatics.}$ \hspace{2mm} Zeynep Akata$^{1,2}$ \hspace{2mm} Bernt Schiele$^{1}$ \hspace{2mm} Andreas Bulling$^{1}$ \vspace{4mm} \\
\begin{tabular}{cc}
$^{1}$Max Planck Institute for Informatics & $^{2}$Amsterdam Machine Learning Lab \\
Saarland Informatics Campus & University of Amsterdam
\end{tabular}
}

\newcommand{\modi}[1]{\textcolor{blue}{#1}}

\maketitle
%\thispagestyle{empty}

%%%%%%%%% ABSTRACT
\begin{abstract}
Zero-shot image classification using auxiliary information, such as attributes describing discriminative object properties, requires time-consuming annotation by domain experts.
We instead propose a method that relies on human gaze as auxiliary information, exploiting that even non-expert users have a natural ability to judge class membership.
We present a data collection paradigm that involves a discrimination task to increase the information content obtained from gaze data.
Our method extracts discriminative descriptors from the data and learns a compatibility function between image and gaze using three novel gaze embeddings: Gaze Histograms (GH), Gaze Features with Grid (GFG) and Gaze Features with Sequence (GFS).
We introduce two new gaze-annotated datasets for fine-grained image classification and show that human gaze data is indeed class discriminative, provides a competitive alternative to expert-annotated attributes, and outperforms other baselines for zero-shot image classification.
\end{abstract}

\section{Introduction}

Zero-shot learning is a challenging task given that some classes are not present at training time~\cite{APHS13,PPH09,RSS11,SGSBMN13}.
State-of-the-art methods rely on auxiliary information to aid the classification, such as object attributes~\cite{FEHF09,FZ07,LNH13}.
While image annotation using such attributes can be performed by na\"ive users, domain experts have to compile the initial list of discriminative attributes for a fixed set of classes and have to revise this list whenever new classes are added.
Several recent works therefore evaluated alternatives, such as distributed text representations extracted from online text corpora such as Wikipedia~\cite{MSCCD13,PSM14}, web-search data~\cite{RSS11} or object hierarchies, such as WordNet~\cite{WordNet}.
While such representations can be extracted automatically and are therefore less costly, they do not outperform attributes.

\begin{figure}[t]
   \centering
    \includegraphics[width=\linewidth]{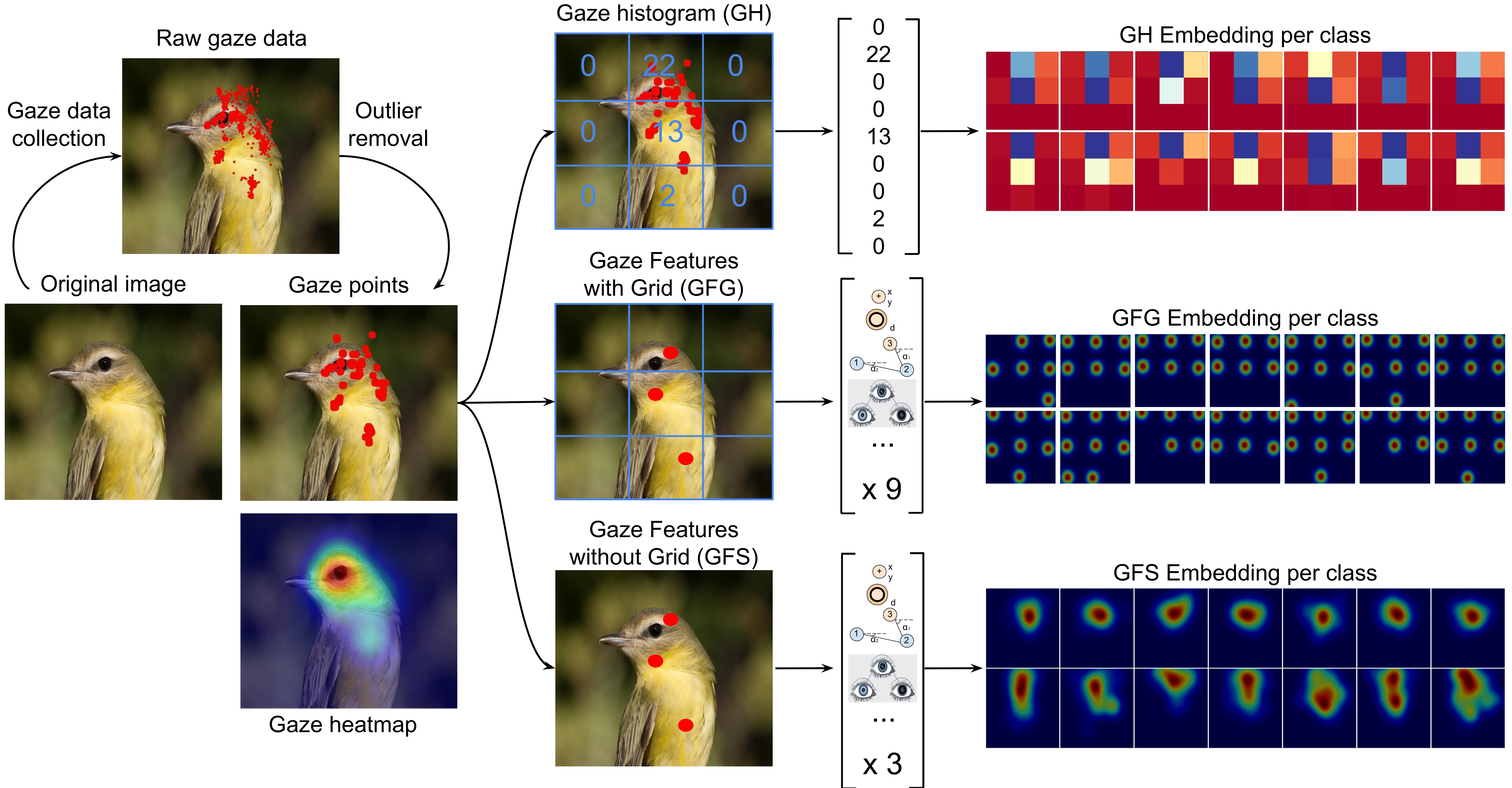}
    %\vspace{-4mm}
\caption{We encode gaze points into vectors using three different methods: gaze histogram (GH), gaze features with grid (GFG), and gaze features with sequence (GFS).}
%\vspace{-4mm}
\label{fig:data_processing}
\end{figure}

We instead propose to exploit human gaze data as auxiliary information for zero-shot image classification.
Gaze has two advantages over attributes: 1) discrimination of objects from different classes can be performed by non-experts, i.e.\ we do not require domain knowledge, and 2) data collection only takes a few seconds per image and is implicit, i.e.\ does not involve explicitly picking class attributes but instead exploits our natural ability to tell objects apart based on their appearance.
We further propose a novel data collection paradigm to encourage observers to focus on most discriminative parts of an object and thereby maximise the information content available for the classification task.
The paradigm involves observers to first inspect exemplars from two different object classes shown to them side-by-side, and subsequently take a binary decision for class membership for another exemplar shown randomly from one of these classes.
While human gaze data has previously been used to obtain bounding-box annotations for object detection~\cite{PCKF14} or approximated by mouse clicks to guide image feature extraction~\cite{DengKrauseFei-Fei_CVPR2013}, this work is first to directly use human gaze data as auxiliary information for zero-shot learning.

The contributions of our work are three-fold. First, we propose human gaze data as auxiliary information for zero-shot image classification, being the first work to tackle this task using gaze.
Second, we provide extensive human gaze data of multiple observers for two fine-grained subsets of Caltech UCSD Birds 2010 (CUB)~\cite{CaltechUCSDBirdsDataset} and one subset of Oxford Pets (PET)~\cite{PVZJ12} datasets.
Third, we propose three novel class-discriminative gaze descriptors, namely Gaze Histograms (GH), Gaze Features with Grid (GFG), and Gaze Features with Sequence (GFS) and complement deep image features in a structured joint embedding framework~\cite{ARWLS15}.
Through extensive evaluations on our datasets, we show that human gaze of non-experts is indeed class-discriminative and that the proposed gaze embedding methods improve over several baselines and provide a competitive alternative to expert-provided attributes for zero-shot learning.

\section{Related Work}
Our work is related to previous works on zero-shot learning and gaze-supported computer vision.

\myparagraph{Zero-Shot Learning.} Zero-shot learning~\cite{APHS13,PPH09,RSS11,SGSBMN13} assumes disjoint sets of training and test classes.
As no labeled visual data is available for some classes during training, typically some form of auxiliary information is used to associate different classes.
Attributes~\cite{FEHF09,FZ07,LNH13} being human-annotated discriminative visual properties of objects are the most common type of auxiliary information. They have been shown to perform well in several tasks such as image classification~\cite{GL11,PG11,SFD11,WF09,WM10}, pedestrian detection~\cite{KBBN09,LSTF12}, and action recognition~\cite{LKS11,QJC11,YJKLGL11}. On the model side, multi-modal joint embedding methods~\cite{APHS13,ARWLS15,XASNHS16} have been shown to provide a means to transfer knowledge from images to classes and vice versa through attributes.
However as fine-grained objects~\cite{OxfordFlowersDataset,CaltechUCSDBirdsDataset} are visually very similar to each other, a large number of attributes are required which is costly to obtain. Therefore, several alternatives have been proposed in the literature. Distributed text representations such as Word2Vec~\cite{MSCCD13} or GloVe~\cite{MSCCD13} are extracted automatically from online textual resources such as Wikipedia.
Hierarchical class embeddings provide another alternative (e.g. using WordNet~\cite{WordNet}) to learn semantic similarities between classes.
On the other hand, search for alternative sources of auxiliary information has introduced the concept of fine-grained visual descriptions~\cite{RALS16} which indicates that although novice users may not know the name of a fine-grained object, they have a natural way of determining discriminative properties of such objects.

Collecting labels from experts or attributes from novice users requires asking many yes/no questions for each image.
We argue that, instead, it may be enough for them to look at an image to identify fine-grained differences between object classes.
Although eye tracking equipment adds to the cost, recent advances suggest that eye tracking will soon become ubiquitous, e.g. in mobile phones~\cite{Huang2017}.
Therefore, we propose to extract class-discriminative representations of human gaze and use them as auxiliary information for zero-shot learning. 

\myparagraph{Gaze-Supported Computer Vision.} Gaze has been an increasingly popular cue to support various computer vision tasks.
Gaze-tracking data has been used to perform weakly supervised training of object detectors~\cite{karthikeyan2013and,Shcherbatyi15_arxiv,yun2013studying}, estimating human pose~\cite{mps13iccv}, inferring scene semantics~\cite{subramanian2011can}, detecting actions~\cite{mathe2014multiple}, detecting salient objects in images~\cite{Li14} and video~\cite{karthikeyan2015eye}, segmenting images~\cite{mishra2009active}, image captioning~\cite{SB16} or predicting search targets during visual search~\cite{sattar15_cvpr}.
Human gaze data is highly dependent on the task the annotators have to complete. While \cite{judd2009learning, yun2013studying} collected gaze tracking data for a free viewing task, \cite{PCKF14} asked users to focus on a visual search task and built POET dataset. On the other hand, in \cite{li2009dataset} gaze has been used to evaluate saliency algorithms on video sequences. \cite{DengKrauseFei-Fei_CVPR2013} imitated human gaze data with ``bubbles'' that they draw around mouse clicks where the annotators find distinguishing image regions. Others used saliency maps instead of real gaze data to improve object detection performance~\cite{ moosmann2006learning,rutishauser2004bottom}.
Maybe the most closely related work to ours is \cite{PCKF14}, where fixations were used to generate object bounding boxes and thereby reduce the bounding-box annotation effort. On the other hand, to the best of our knowledge, we are first to collect real eye tracking data to extract class-discriminative representations and then in turn to use them as auxiliary information for the specific task of zero-shot image classification. Our technical novelty is in our design of effective gaze representations that provide a structure in class embedding space.

\section{Gaze Tracking and Datasets}
Here, we present our gaze data collection paradigm, detail our gaze datasets and our gaze embeddings.

\subsection{Gaze Data Collection}
\label{subsec:data_coll}

We collect the eye tracking data with the Tobii TX300 remote eye tracker that records binocular gaze data at 300 Hz.
We implement a custom data collection software in C$\#$ using the manufacturer-provided SDK which we will make publicly available.
Our software logs a timestamp, users' on-screen gaze location, their pupil diameter, as well as a validity code for each eye that indicates the tracker’s confidence in correctly identifying the eye. We use gaze-points that are valid for both eyes. The participants are seated $67$ cm from a $31.5$ inch LCD screen. We use a chin rest to reduce head movement and consequently improve the eye tracking accuracy. The vertical extend of the image shown on the screen is $\approx 15$ cm, thus the visual angle\footnote{Visual Angle $= 2 \times \arctan$ (Vertical Extend Stimuli / Distance)} is $\approx \ang{25}$.
We record 5 participants for every image which leads to 5 gaze streams for each image of three datasets.
Almost $50\%$ of our participants have impaired eye sight however, none of them wear glasses during the data collection although $30\%$ of them wear contact lenses. 

\begin{figure}[t]
   \centering
    \includegraphics[width=\linewidth, trim=0 50 0 0, clip]{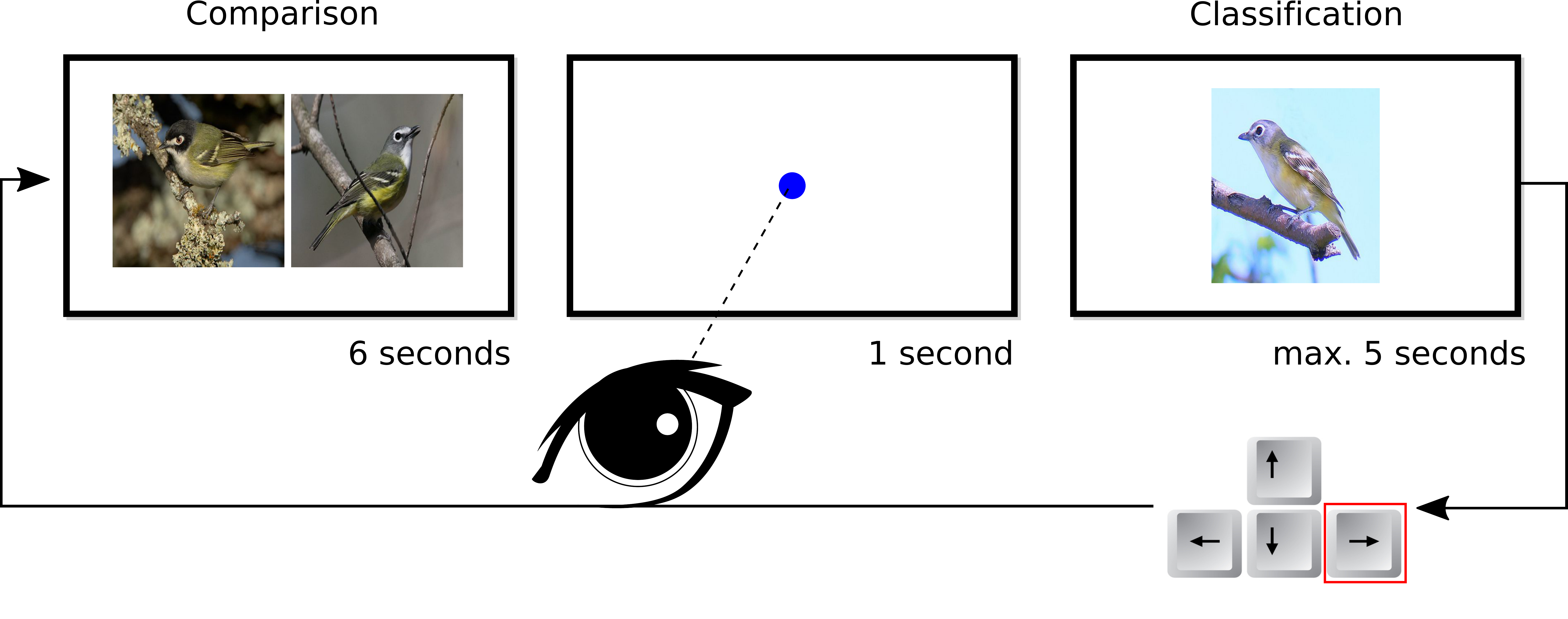}
    %\vspace{-3mm}
\caption{Participants first look at two images of two fine-grained classes (6 sec), then at the center of the screen to ``reset'' their gaze position (1 sec), finally they click the left or right arrow on the keyboard (max 5 sec) to select the class they think the image belongs to. We record their gaze only on classification screen.}
%\vspace{-3mm}
\label{fig:data_collection}
\end{figure}

Our data collection paradigm is illustrated in~\autoref{fig:data_collection}.
Participants first answer a short questionnaire on demographics, e.g. age, gender, eye sight, etc. and then we calibrate the eye tracker using the standard 5-point calibration routine. 
After calibration, participants follow a cycle of three steps, namely comparison, 
fixation, and classification.
During the comparison step, we show two 
example images that we randomly sample from two fine-grained classes for six seconds where participants learn fine-grained differences between two classes.
In the fixation step, we ask the participants to fixate on a dot in the center of the screen for one second to ``reset'' the gaze position.
In the classification step, we show a new instance of one of the two classes which participants need to classify by clicking on right/left arrow of the keyboard in max five seconds. This step terminates before five seconds if the annotator decides earlier.
A new cycle starts until all the images are annotated by the same user.

\myparagraph{Gaze Datasets.}
We collect gaze tracking data for images of two publicly available datasets (See \autoref{tab:datasets} for details).
Following~\cite{DengKrauseFei-Fei_CVPR2013} we collect gaze tracking data of 14 classes ($7$ classes of Vireos and $7$ classes of Woodpeckers: CUB-VW) for all the available $464$ images. Each image is annotated by 5 participants. 
In addition, CUB-VWSW includes two more bird families of Sparrows and Warblers. CUB-VWSW contains $11,730$ gaze tracks of five participants for every image, i.e. $1882$ images in total.
Finally, we collect gaze tracking data for the images of a $24$-class subset of the Oxford Pets dataset~\cite{PVZJ12}, where we take all $12$ classes from Cats and a subset of $12$ classes from Dogs. Following CUB setting, we collect $3,600$ gaze tracks of $720$ images from five participants and name this dataset PET. We collect gaze data at the sub-species level, e.g. black-capped vireo vs red-eyed vireo. We observe that comparing birds at a higher level, e.g. woodpeckers vs vireos, is too easy and users take a decision instantly, while comparing birds at the sub-species level takes longer providing us more gaze points.

\subsection{Gaze Embeddings}
\label{subsec:data_proc}
We propose Gaze Histograms (GH), Gaze Features with Grid (GFG) and Gaze Features with Sequence (GFS) as three gaze embedding methods.

\myparagraph{Gaze Histogram (GH).} Gaze points are encoded into a $m \times n$-dimensional vector using a spatial grid overlayed over the image with $m$ rows and $n$ columns.
The per-class gaze histogram embedding is the mean gaze histogram of a particular class.
We encode gaze of each participant separately to evaluate the annotator bias.

\autoref{fig:data_processing} (top) shows how we construct a 9-dimensional histogram using a spatial grid of $3\times3$. For simplicity, we show per-class histograms of seven vireos and seven woodpeckers with darker colors indicating higher number of occurrences.
High attention points for Vireos (top row) fall in the middle of the image whereas for woodpeckers (bottom row) the top of the image seems to be more important.
From visual inspection of the original images, we observe that vireos in CUB often sit on horizontal tree branches with their eyes being the most discriminative property.
In contrast, woodpeckers often climb on large tree trunks with their head region being the most discriminative property.
Gaze histograms capture these spatial, \ie horizontal versus vertical, and class-specific differences.

\begin{figure}[t]
   \centering
    \includegraphics[width=\linewidth]{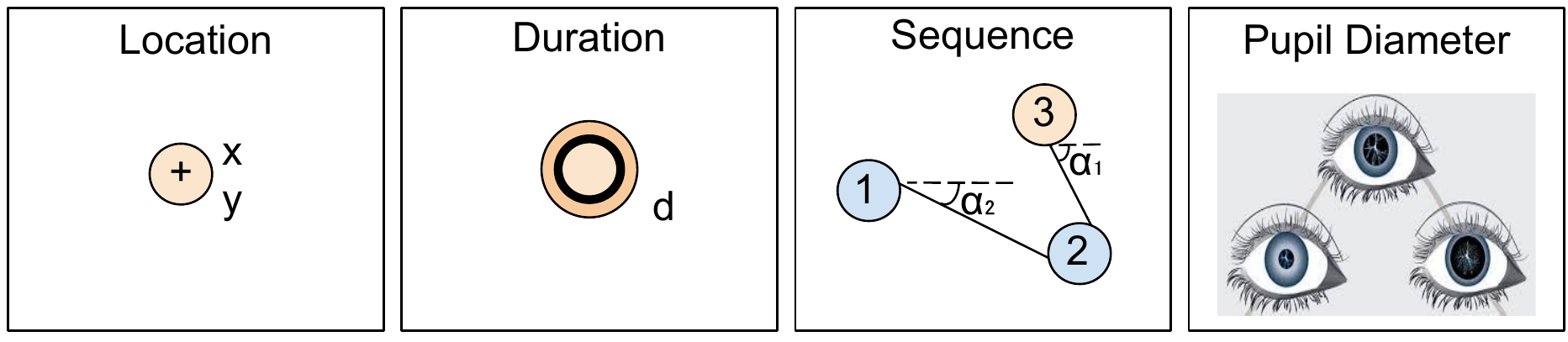}
    %\vspace{-3mm}
\caption{Gaze features include gaze point location ($x,y$), gaze point duration ($d$), angles to the previous and next gaze points in the sequence ($\alpha_1,\alpha_2$), and the pupil diameter ($R$).}
%\vspace{-3mm}
\label{fig:gaze_parameters}
\end{figure}

\myparagraph{Gaze Features (GFx).} Counting the number of gaze points that fall into a grid cell encodes the location information coarsely and does not encode any information about the  duration, sequence and the attention of the observer. Therefore, we build 6-dimensional gaze features, i.e. $[x,y,d,\alpha_1,\alpha_2,R]$, as shown in  \autoref{fig:gaze_parameters}.
Our gaze features encode gaze location ($x,y$), gaze duration ($d$), angles ($\alpha_1,\alpha_2$) between the previous and subsequent gaze point in the scan path, and pupil diameter $R$ that was shown to relate to processing load~\cite{hyona1995pupil} of the observer. We embed these gaze features in two different ways, namely Gaze Features with Grid (GFG) and Gaze Features with Sequence (GFS).

\emph{Gaze Features with Grid (GFG)} uses a spatial grid similar to gaze histograms (GH) to discretize the gaze space.
Instead of counting the number of gaze-points per cell, we average the 6-dim gaze features of the points that fall in each cell. We then concatenate gaze features that fall inside each grid cell into a $6 \times m \times n$-dimensional vector with $m$ and $n$ being the number of rows and columns of the spatial grid.
The per-class GFG embedding is then the average of all GFG vectors from the same class. By encoding the spatial ordering of the gaze points, GFG captures information related to the typical behavioral patterns of birds such as sitting on horizontal tree branches vs climbing large tree trunks. 

\emph{Gaze Features with Sequence (GFS)} encodes the sequential order of gaze points.
First we order gaze points with respect to time, i.e. first occurring gaze point to the last, then we sequentially select a fixed number ($k$) of gaze points from each gaze sequence and embed them as a $6 \times k$-dimensional vector. Here, $k$ is typically the minimum number of gaze points extracted from the gaze-sequence of a certain observer. The GFS encodes the time sequence of the gaze points instead of focusing on their spatial layout. The per-class GFS embedding is the average GFS embeddings of the same class.

\myparagraph{Combining Gaze Embeddings.} As participants gaze at different regions of the same image, we argue that their gaze embeddings may contain complementary information.
We thus propose three different methods to combine their gaze embeddings:
First, we average the per-class gaze embeddings ($\phi(y)$) of each participant abbreviated by \texttt{AVG}. Second, we concatenate per-class gaze embedding of each participant through early fusion, \ie \texttt{EARLY} and learn one single model. Third, we learn a model for each participant separately and then we average their classification scores before making the final prediction decision in the late fusion setting, \ie \texttt{LATE}.

\section{Gaze-Supported Zero-Shot Learning}
In zero-shot learning, the set of training and test classes are disjoint. During training, the models have access to only the images and gaze embeddings of training classes but none of the images or gaze embeddings of test classes. The lack of labeled images from test classes is compensated by the use of auxiliary information that defines a structure in the label space~\cite{APHS15,ARWLS15,GL11,WBU11} and provides a means of associating training and test classes. In the following, we provide the details of the zero-shot learning model~\cite{ARWLS15}.

\myparagraph{Zero-Shot Learning Model.}
Given image and class pairs $x_n \in \X$ and $y_n \in \Y$ from a training set ${\cal S}=\{(x_n,y_n), n =1 \ldots N\}$, we use the Structured Joint Embedding (SJE) model~\cite{ARWLS15} to learn a function $f: \X \rightarrow \Y$ by minimizing the empirical risk
\begin{equation}
 \frac{1}{N} \sum_{n=1}^N \D(y_n,f(x_n))
\end{equation}
where $\D: \Y \times \Y \rightarrow \{0,1\}$ defines the cost of predicting $f(x)$ when the true label is $y$. 
The SJE model maximizes the compatibility function $F: \X \times \Y \rightarrow \Re$ as follows:
\begin{equation}
f(x;W) = \arg \max_{y \in \Y} F(x,y; W).
\label{eq:annot}
\end{equation}
that has the following bi-linear form:
\begin{equation}
F(x,y;W) = \theta(x)^{\top} W \p(y) . 
\label{eqn:form}
\end{equation}
where the image embedding ($\theta(x)$), \ie image features extracted from a Deep Neural Network (DNN) and the class embedding ($\p(y)$), \ie gaze embeddings are provided as a pre-processing step. $W$ is learned through structured SVM~\cite{TJH05} by maximizing the ranking of the correct label:
\begin{equation}
\max_y(\Delta(y_n,y) + F(x_n,y;W)) - F(x_n,y_n;W)
\end{equation}
and optimized through stochastic gradient descent (SGD). At test time, we search for the test class whose per-class gaze embedding yields the highest joint compatibility score.

\section{Experiments}
\label{sec:exp}

{
\renewcommand{\arraystretch}{1.2}
\begin{table}[t]
 \begin{center}
  \begin{tabular}{lccc}
    \textbf{Dataset} & \textbf{\# img / class} & \textbf{Gaze}  & \textbf{Bubbles~\cite{DengKrauseFei-Fei_CVPR2013}}\\     \hline
    CUB-VW & 464 / 14 & 2320 & 210\\
    CUB-VWSW & 2346 / 60 & 11730 & 900 \\
    PET & 720 / 24 & 3600 & --
  \end{tabular}
 \end{center}
 \vspace{-3mm}
\caption{Statistics for CUB-VW, CUB-VWSW datasets (images selected from CUB~\cite{CaltechUCSDBirdsDataset}) and PET dataset (images selected from Oxford PET~\cite{PVZJ12}) w.r.t. number of images, classes, number of gaze and bubble~\cite{DengKrauseFei-Fei_CVPR2013} tracks.}
\vspace{-3mm}
\label{tab:datasets}
\end{table}
}

In this section, we first provide details on datasets, image embeddings and parameter setting that we use for zero-shot learning. We then present our detailed evaluation of gaze embeddings compared with various baselines both qualitatively and quantitatively.

\myparagraph{Datasets.} As shown in~\autoref{tab:datasets},
\cite{DengKrauseFei-Fei_CVPR2013} provide mouse-click data, i.e. bubble tracks, for 14 classes ($7$ classes of Vireos and $7$ classes of Woodpeckers: CUB-VW) of CUB for a selection of $210$ images. 
They collected bubble tracks for every image, however every annotator did not annotate every image. Therefore, unlike our 5 streams of gaze-tracks collected from 5 participants, there is only a single stream of bubble-tracks. On CUB-VW bubble-tracks, we build per-class bubble representations in three different ways, i.e. the same as gaze, and found out that Bubble Features with Sequence (BFS), encoding $x,y$ location and radius of the bubble works the best, therefore we use these as bubble representations in all our experiments. We extensively evaluate our method on CUB-VW in the following section. Note that we validate all the gaze-data processing parameters on CUB-VW and use the same parameters for other datasets. 
Our CUB-VWSW dataset, i.e. including Vireos, Woodpeckers, Sparrows and Warblers, comes with $312$ expert-annotated attributes for every class, i.e. embedded as $312$-dimensional per-class attribute vectors. \cite{DKSF16} extended \cite{DengKrauseFei-Fei_CVPR2013} by collecting bubble-tracks for more bird species, therefore bubble-tracks of $900$ images that~\cite{DKSF16} selected for CUB-VWSW dataset are also available. 
The PET dataset neither contains attributes nor bubble-tracks. 
For both our CUB-VWSV and PET datasets, we further construct bag-of-words representations extracted from Wikipedia articles that describe a specific class to build per-class representations. Bag of words frequencies are produced by counting the occurrence of each vocabulary word that appears within a document. To obtain fixed-sized descriptors, we only consider the N-most frequent words across all classes after removing stop-words and stemming.

\myparagraph{Image Embeddings and Parameter Setting.} As image embeddings, we extract $1,024$-dim CNN features from an ImageNet pre-trained GoogLeNet~\cite{szegedy2014going} model. We neither do any task-specific image processing, such as image cropping, nor fine-tune the pre-trained network on our task.
We cross-validate the zero-shot learning parameters, i.e. step size in SGD and the number of epochs, on 10 different zero-shot splits that construct by maintaining a ratio of $2/1/1$ for disjoint training, validation and test classes. We measure accuracy as average per-class top-1 accuracy.

\subsection{Gaze Embeddings on CUB-VW}
\label{subsec:baselines}
In this section, we first show how we pre-process our raw gaze data, and then extensively evaluate our gaze embeddings wrt. multiple criteria on the CUB-VW dataset.

\begin{figure}[t]
  \centering   
    \includegraphics[width=0.49\linewidth, trim=10 0 70 0,clip]{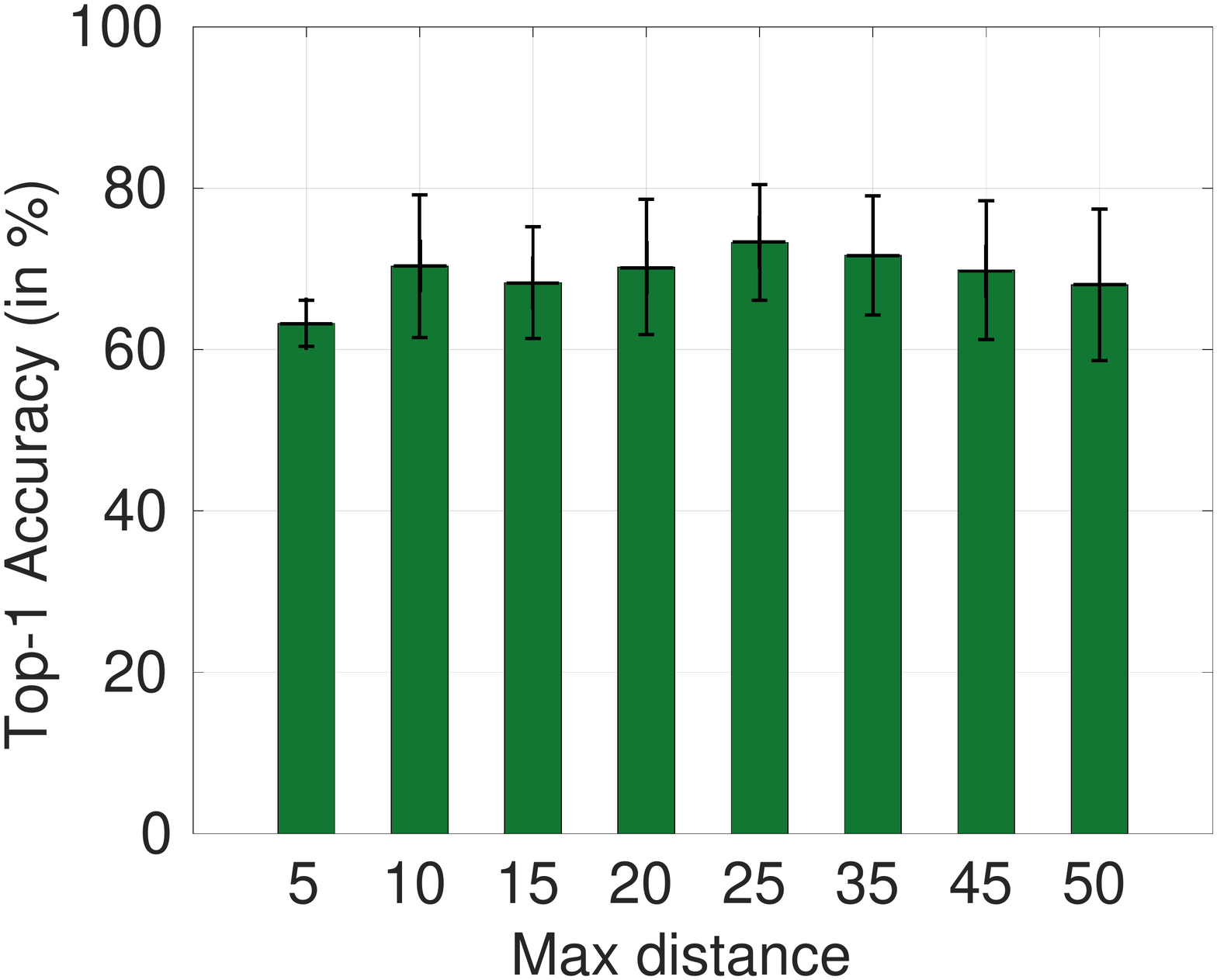}
    \includegraphics[width=0.49\linewidth, trim=10 0 70 0,clip]{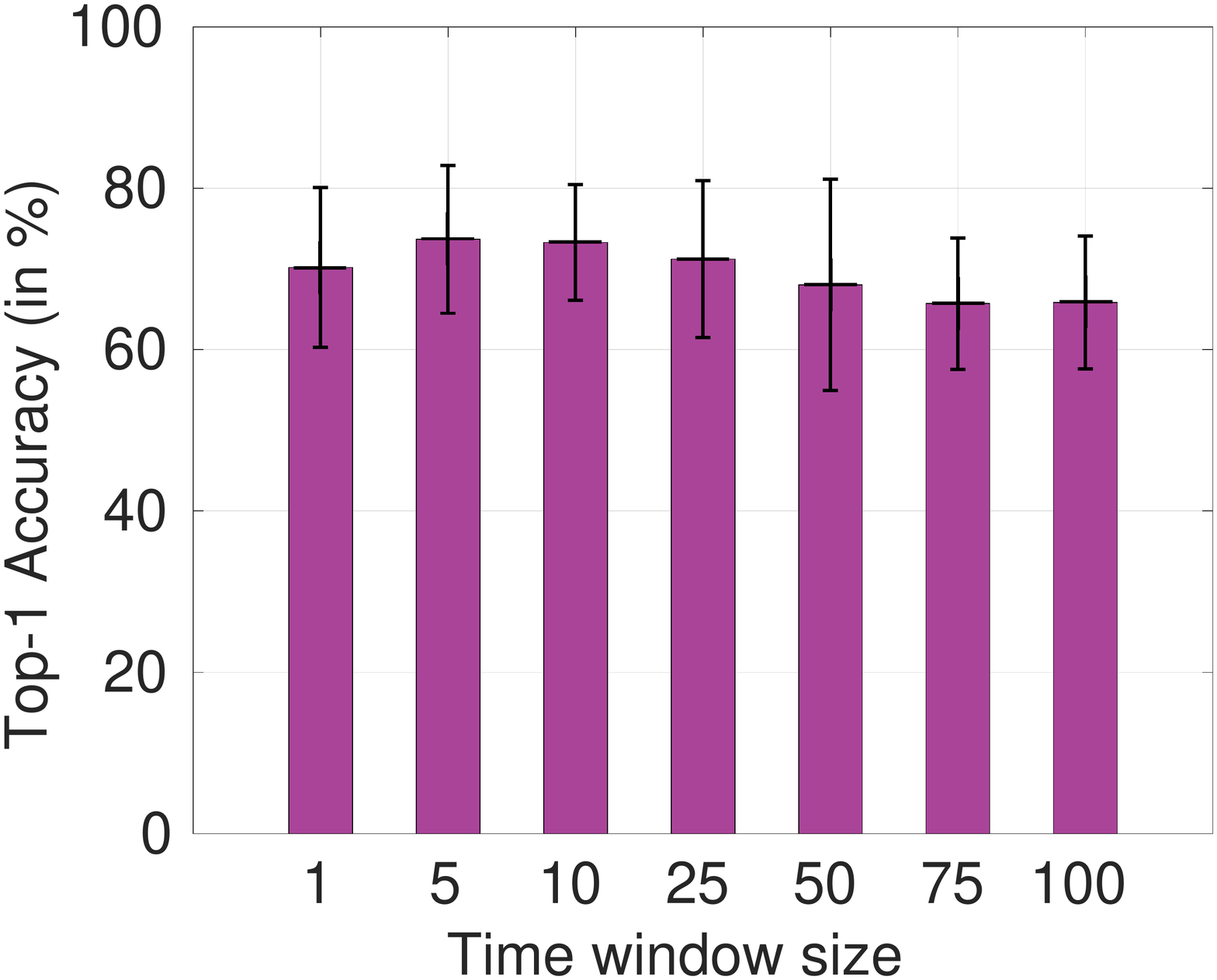}
    %\vspace{-3mm}
  \caption{Raw gaze data processing: max distance between gaze points and time window size.}
  \vspace{-3mm}
\label{fig:gaze_param}
\end{figure}

\myparagraph{Processing Raw Gaze Data.} 
Raw gaze data is inherently noisy due to inaccuracies of the eye tracker.
We reduce this noise using a dispersion-based method~\cite{salvucci2000identifying} which calculates the dispersion of gaze points using a sliding window approach with window size $w_s$ and applies a threshold $t_s$ on this dispersion value.
All gaze points within the window are then set to the mean of all points below the threshold. 
In order to disentangle this raw-data pre-processing step from our end task of zero-shot learning, we train 
standard one-vs-rest SVM classifiers on stacked gaze features as training samples, and image label as supervision signal. We use $[x, y, d, \alpha_1, \alpha_2, R]$ as gaze features, GFS as gaze feature encoding and evaluate 10 random train and test splits.

\autoref{fig:gaze_param} (left) shows that the highest accuracy is obtained using $w_s = 25$ degrees (among $w_s = 5 ... 50$).
Time-window size ($t_s$) depends on how long the annotator needs to view the image before making a decision.
As our users have significantly shorter viewing duration ($\approx 0.5$sec) compared to eye tracking studies that requires long viewing times, e.g. reading a textual document, we fix time window size on our data.
By keeping $w_s = 25$, we evaluate $t_s = 1 ... 100$ and observe that $t_s=10$ms works the best (\autoref{fig:gaze_param}, right). 
We observe that performance does vary across experiments, albeit not significantly.
Therefore, at least for the datasets investigated in this work, gaze data can be processed in a generic fashion, i.e.\ does not have to be tailored for a particular user or object class.
\begin{figure}[t]
    \begin{center}
\includegraphics[width=0.49\linewidth, trim=10 0 70 0,clip]{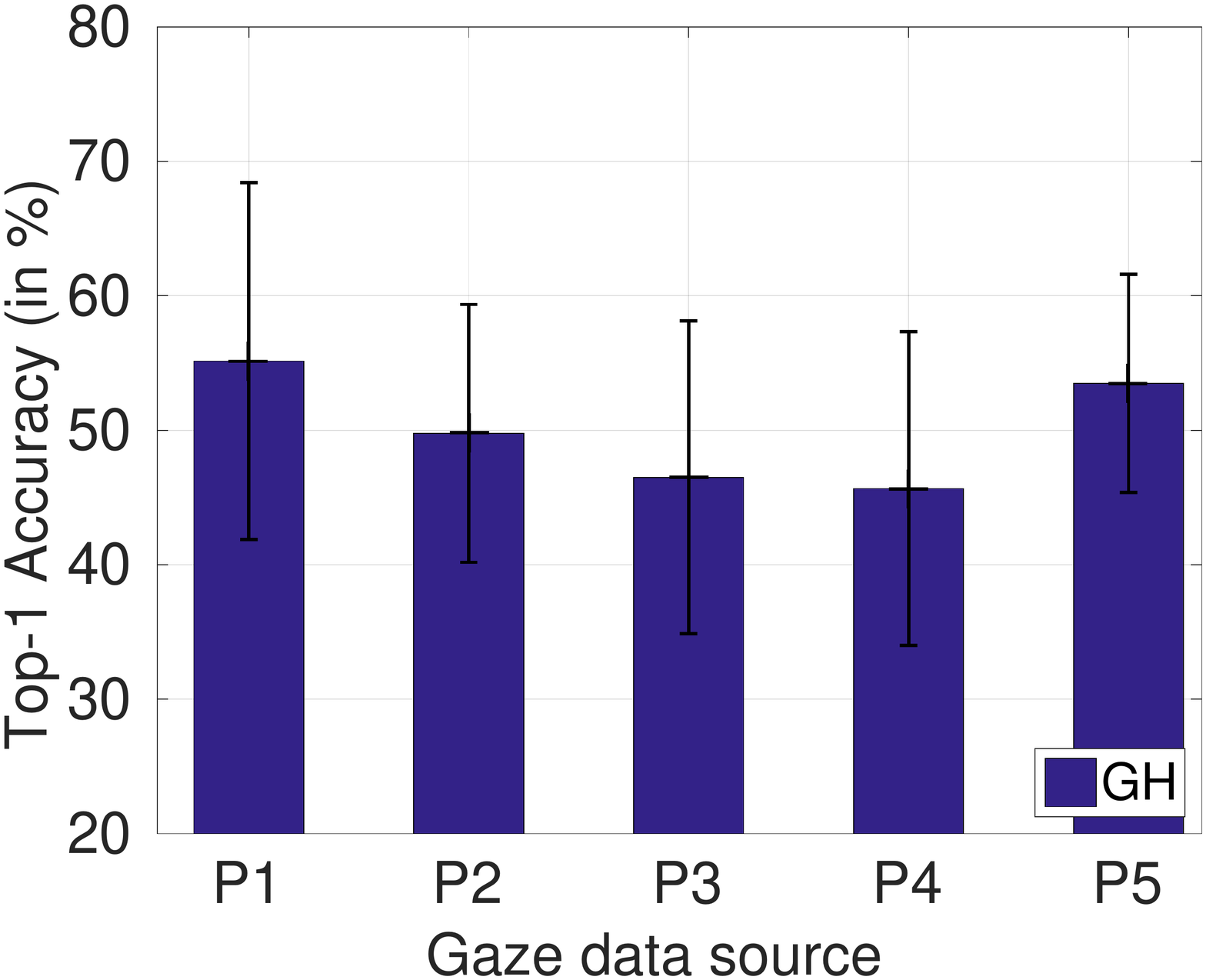} 
\includegraphics[width=0.49\linewidth, trim=10 0 70 0,clip]{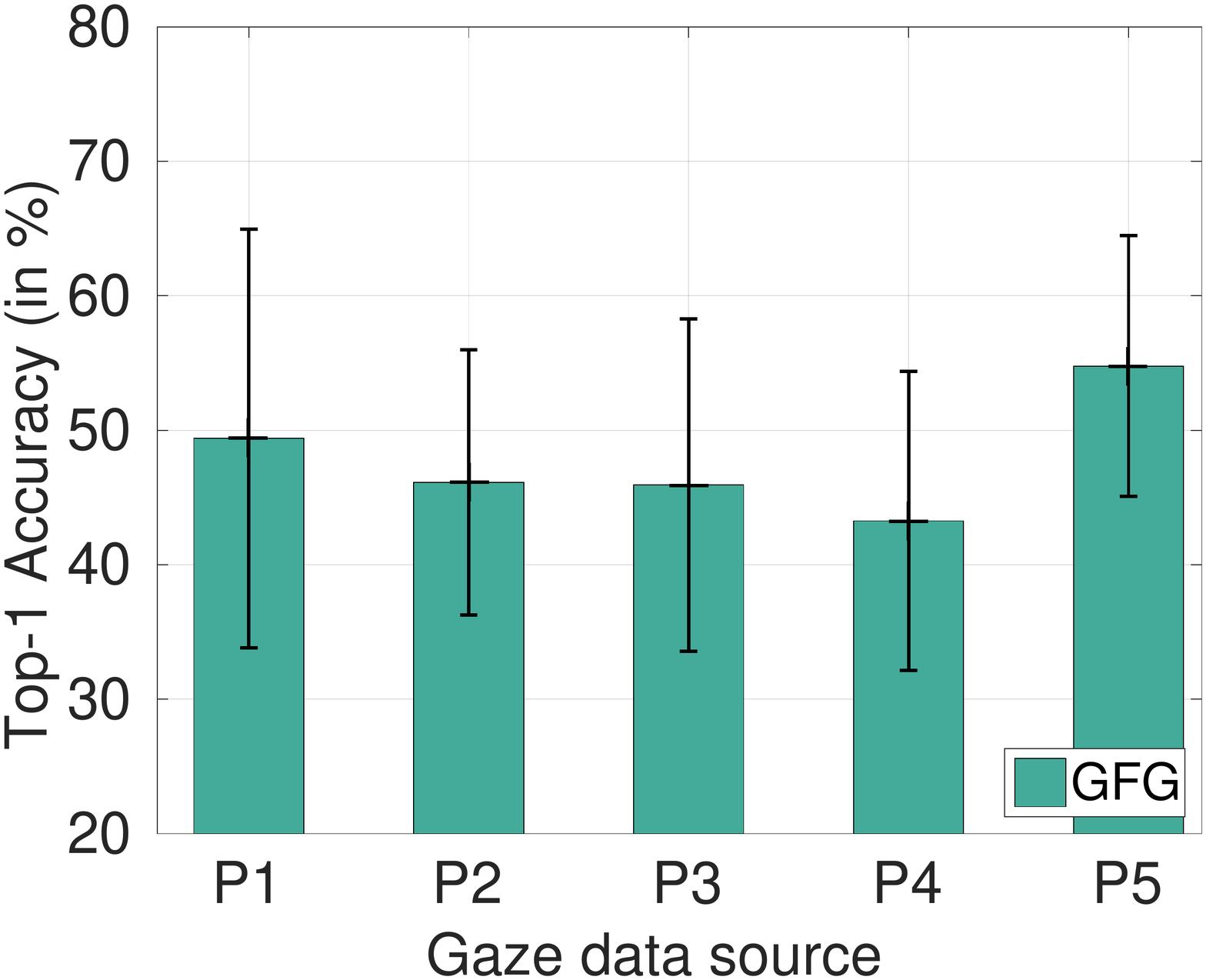} \\
\includegraphics[width=0.49\linewidth, trim=10 0 70 0,clip]{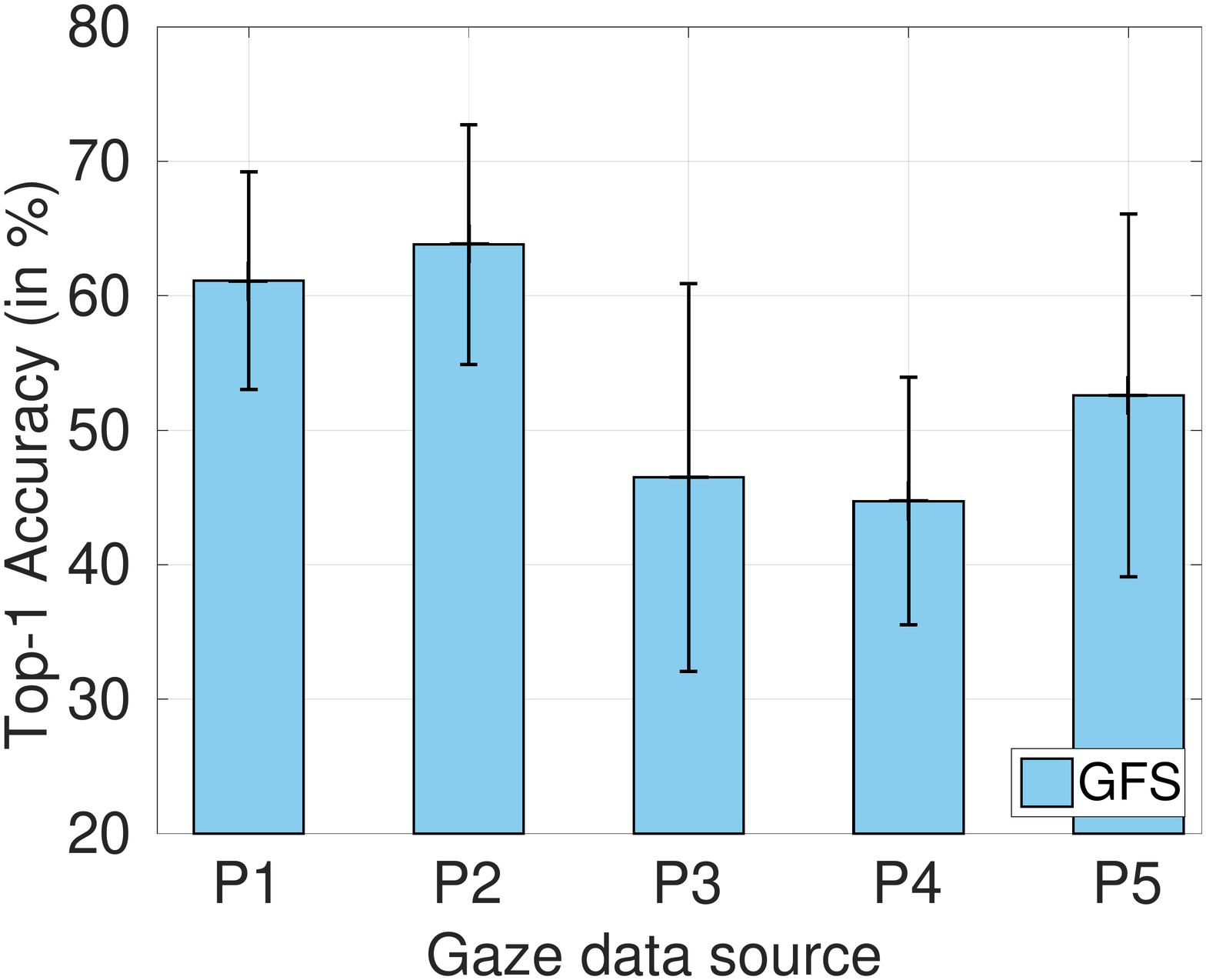}
\includegraphics[width=0.49\linewidth, trim=10 0 70 0,clip]{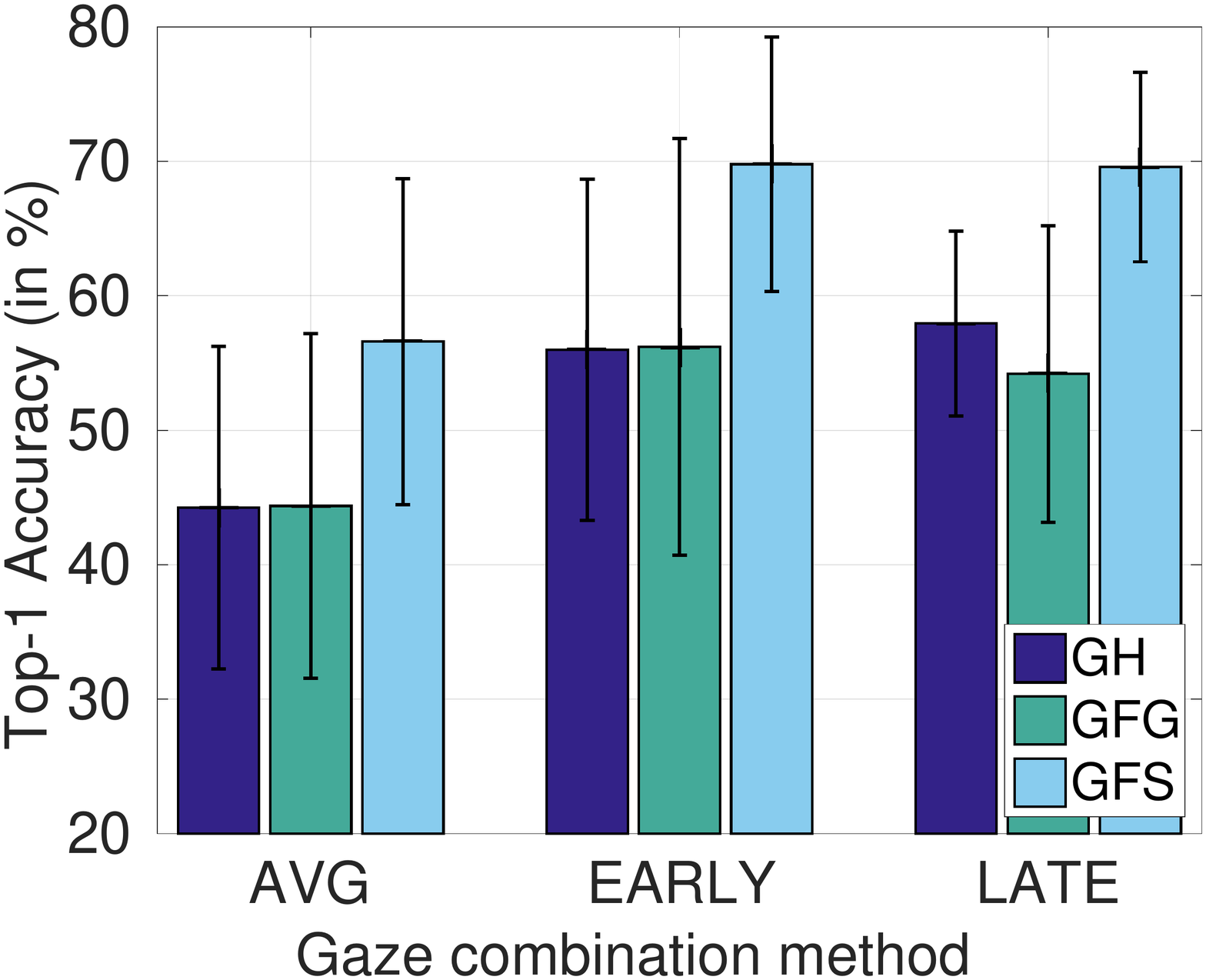}
  \end{center}
  \vspace{-4mm}
\caption{Comparing Gaze Histogram (GH), Gaze Features with Grid (GFG) and Gaze Features with Sequence (GFS). We evaluate 5 participants separately as well as their various combinations: Averaging each participant's  gaze embeddings (AVG), Combining them through early fusion (EARLY) and through late fusion (LATE).}
\vspace{-3mm}
\label{fig:cub_att}
\end{figure}

\myparagraph{Comparing Different Gaze Embeddings.} We now compare the performance of Gaze Histograms (GH), Gaze Features with Grid (GFG) and Gaze Features with Sequence (GFS). We build GFx, i.e. GFG and GFS, with all gaze features, i.e. $[x,y,d,\alpha,R]$ for consistency.
We first consider the gaze embeddings of our 5 participants separately and then combine the gaze embeddings of each participant by averaging them (\texttt{AVG}), concatenating them through early fusion (\texttt{EARLY}), and combining the classification scores obtained by each participant's gaze data through late fusion (\texttt{LATE}). We repeat these experiments on 10 different zero-shot splits to show a robustness estimate. 

As shown in \autoref{fig:cub_att}, GFS embeddings outperform GH and GFG embeddings.
This implies that the sequence information is more helpful than the spatial discretization by using the grid. Therefore, we argue that in fine-grained zero-shot learning task, the sequence of gaze points is important to obtain best performance.
Our second observation is that indeed each participant's gaze embedding lead to different results, therefore considering the annotator bias is important. For GH, the best performing participant is the first, while for GFG it is the fifth and for GFS it is the second. We argue that gaze embeddings of each participant is complementary, therefore we propose to combine gaze embeddings of different participants. We obtain $56.6\%$ using \texttt{AVG}, $69.8\%$ using \texttt{EARLY} and $69.6\%$ with \texttt{LATE}. These results support our intuition that there is complementary information between gaze embeddings of different participants. 

\begin{figure}[t]
\begin{center}
\includegraphics[width=0.9\linewidth, trim=30 5 70 30,clip]{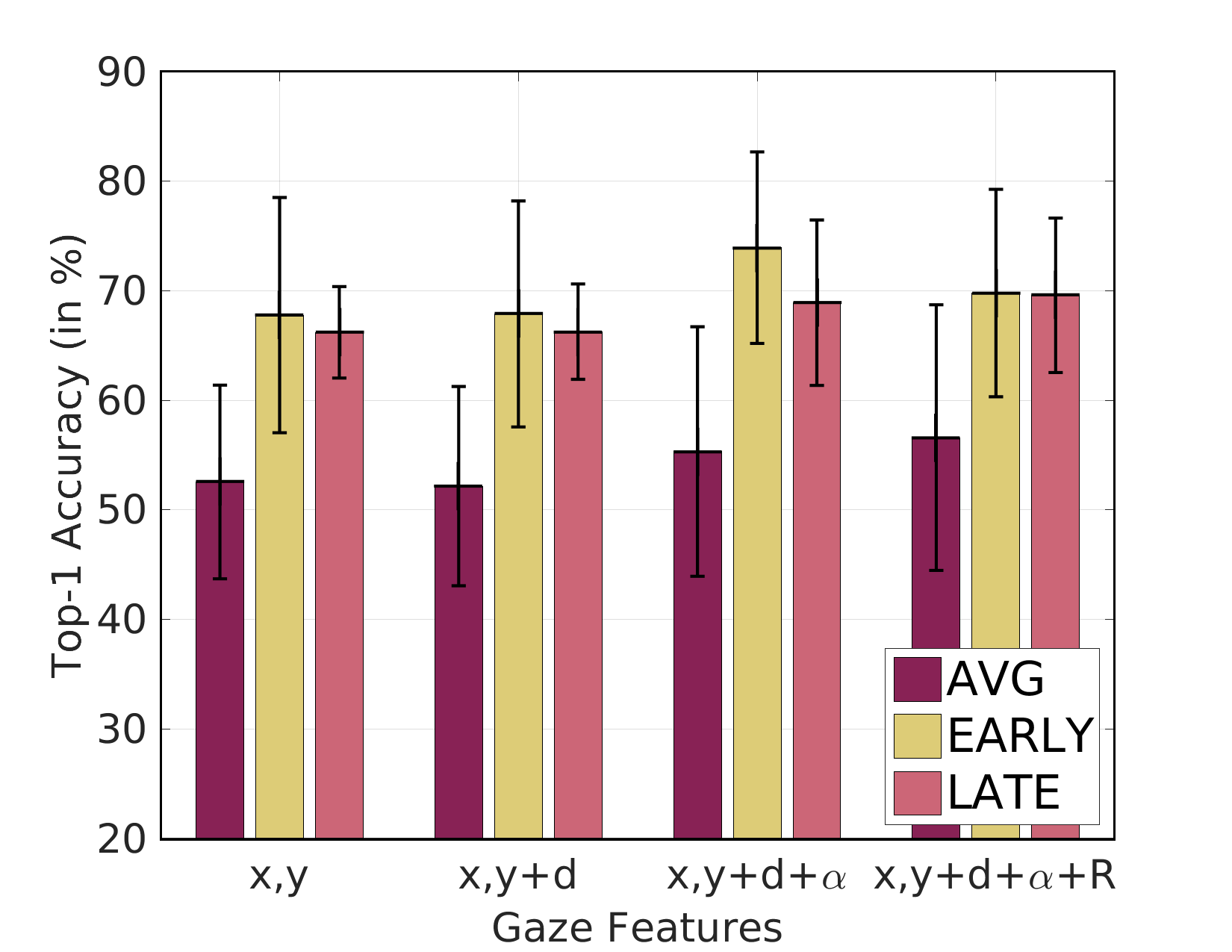}%{gaze_features_new}
\end{center}
\vspace{-4mm}
\caption{Effect of gaze features: Location ($x,y$), Duration (d), Sequence ($\alpha_1, \alpha_2$), Pupil diameter (R). We start with $x,y$ and concatenate d, $\alpha_1, \alpha_2$ and R cumulatively.}
\vspace{-3mm}
\label{fig:CUB_gazeparams}
\end{figure}

\myparagraph{Analyzing Gaze Features.}
We evaluate the effects of encoding gaze location ($x,y$), duration (d), sequence ($\alpha_1, \alpha_2$) and the annotator's pupil diameter (R) that measures concentration. 
We build gaze features cumulatively by starting with the parameter $x,y$, followed by d, $\alpha_1, \alpha_2$ and $R$ sequentially in this order. Being the best performing method from \autoref{fig:cub_att}, we evaluate the effect of gaze features with respect to participants combined GFS with \texttt{AVG}, \texttt{EARLY} and \texttt{LATE}. 
We observe from~\autoref{fig:CUB_gazeparams} that \texttt{EARLY} achieves the highest $73.9\%$ accuracy when we use $[x,y,d,\alpha_1,\alpha_2]$ features. The $[x,y]$ features already achieves high accuracy, adding duration, i.e. $d$ slightly improves results and adding the sequence information, i.e. $\alpha$, adds further improvement. However, the pupil diameter parameter does not bring further improvements. As our annotators go through all the images which requires a total of one hour of constant concentration. Although they take a break after half an hour, their concentration drops towards the end of the task while they become familiar with the fine-grained bird species.

{
\renewcommand{\arraystretch}{1.2}
\begin{table}[t]
 \begin{center}
  \begin{tabular}{llc}
    \multicolumn{2}{c}{\textbf{Method}} & \textbf{Accuracy} \\     \hline
    \multirow{5}{*}{Baselines}& Saliency histogram  & $35.8$ \\
    			& Random points in the image & $39.5$ \\
    			& Central gaze point 	& $41.5$ \\	
    			& Bubbles~\cite{DengKrauseFei-Fei_CVPR2013} & $43.2$ \\ 
    			& Bag-of-Words from Wiki		& $55.2$ \\			\hline
   	SoA 		& Human annotated attributes 	& $72.9$ \\			\hline
   	\multirow{2}{*}{Ours} & Gaze embeddings 	& $73.9$ \\ 		    
    			& Attributes + Gaze & $\textbf{78.2}$ 
  \end{tabular}
 \end{center}
 \vspace{-5mm}
\caption{Comparing random points, mean gaze point, saliency histogram using~\cite{GBVS}, bubbles ~\cite{DengKrauseFei-Fei_CVPR2013}, Bag of Words and expert annotated attributes on CUB-VW.}
\vspace{-3mm}
\label{tab:cub14_baseComp}
\end{table}
}
\myparagraph{Comparing Gaze and Baselines.} \autoref{tab:cub14_baseComp} shows a performance comparison of our gaze embeddings with several baselines.
Saliency histogram ($35.8\%$) is a discretization of a saliency map~\cite{GBVS} using a spatial grid over the image.
As a second baseline, we randomly sample points in the image and obtain $39.5\%$ accuracy. Another baseline is taking the location of the central point as an embedding, which leads to $41.5\%$, indicating a certain center bias in CUB-VW images.
Bubbles~\cite{DengKrauseFei-Fei_CVPR2013}, mouse-click locations of visually distinguishing object properties, are the closest alternative to our gaze data.
Bubbles achieve $43.2\%$ accuracy, which supports the hypothesis that non-expert users are able to determine distinguishing properties of fine-grained objects. As the final baseline, we evaluate class embeddings extracted from Wikipedia articles ($55.2\%$). 
Our best performing gaze embeddings, i.e. GFS \texttt{EARLY} with $[x,y,d,\alpha]$ from~\autoref{fig:CUB_gazeparams},  achieve $73.9\%$ accuracy and outperform all these baselines. Moreover, they outperform expert annotated attributes with $72.9\%$ being the current state-of-the-art. This result shows that human gaze data is indeed class discriminative while being more efficient than attributes to collect. Finally, we combine our gaze embeddings with attributes and show by obtaining $78.2\%$ accuracy that human gaze data contains complementary information to attributes.

{
\renewcommand{\arraystretch}{1.2}
\begin{table}[t]
   \centering 
   \begin{tabular}{lc}

    \textbf{Method} & \textbf{Accuracy} \\   \hline
    Gaze & $\mathbf{73.9}$  \\	
    Gaze: same images as bubbles & $69.7$	   \\    
    Gaze: same location as bubbles & $64.0$	   \\   
    Gaze: same number as bubbles (avg)	& $55.0$  	   \\  
    Gaze: same number as bubbles (rnd)	& $49.2$  	   \\   \hline
    Bubbles (mouse-clicks)	& $43.2$
    \end{tabular}
    \vspace{-2mm}
	\caption{Ablation from gaze to bubbles: using our full gaze data with GFS \texttt{EARLY} embedding, using same images as bubbles, concatenating gaze points located inside bubbles, averaging those gaze points and using one among those gaze points vs bubbles.} 
\label{fig:ablation}
\vspace{-3mm}
\end{table} 
}
\myparagraph{Ablation from Gaze to Bubbles.} As we observed a large accuracy gap between gaze and bubble embeddings previously, we now investigate the reason for this gap through an ablation study. We gradually decrease the information content of gaze embeddings in the following way. We first use the same images as bubbles and observe from \autoref{fig:ablation} that the accuracy decreases from $73.9\%$ to $69.7\%$. We then concatenate the gaze features of the gaze points that fall inside the bubbles, i.e. use gaze points at the same location as bubbles, and observe the accuracy decline to $64.0\%$. Instead of concatenating, averaging the gaze points or taking one random point inside bubbles decreases the accuracy to $55.0\%$ and $49.2\%$ respectively. We attribute the accuracy difference between $49.2\%$ and $43.2$ (bubbles) to the gaze features, i.e. $[x,y,d,\alpha,R]$. We conclude from this experiment that the images that the annotators viewed while we recorded their gaze as well as their attention and the quantity, the location, the duration of the gaze-tracks are all important to obtain good zero-shot learning results.

{
 \setlength{\tabcolsep}{6pt}
\renewcommand{\arraystretch}{1.2}
\begin{table}[t]
 \centering
   \begin{tabular}{ll ccc}
      & & \multicolumn{2}{c}{\textbf{CUB}} &  \\
     \textbf{Method} &\textbf{Side-Info}& \textbf{VW} & \textbf{VWSW} & \textbf{PET} \\
     \hline
     Random points & Image & $39.5$ & $9.0$ & $21.0$ \\ 
     Bubbles & Novice & $43.2$ & $10.3$ & N/A \\ 
     Bag of Words & Wikipedia & $55.2$ & $24.0$ & $33.5$ \\
     Human Gaze & Novice & $\mathbf{73.9}$ & $26.0$ & $\mathbf{46.6}$ \\	
     Attributes & Expert & $72.9$ & $\mathbf{42.7}$ & N/A 
   \end{tabular} 
   \vspace{-2mm}
\caption{Comparing random points, bubbles~\cite{DengKrauseFei-Fei_CVPR2013}, bag of words, attributes, and our gaze embeddings (GFS \texttt{EARLY}), on CUB-VW = CUB with Vireos and Woodpeckers, CUB-VWSW = CUB with Vireos, Warblers, Sparrows, Woodpeckers and PET=Oxford Pets with Cats and Dogs.}
%\vspace{-3mm}
\label{tab:others}
\end{table}
}

\subsection{Gaze Embeddings on Other Datasets}
\label{subsec:other}
In this section, we first evaluate gaze embeddings on CUB with 60 species of Vireos, Woodpeckers, Sparrows and Warblers (CUB-VWSW)~\cite{CaltechUCSDBirdsDataset}. To show the generalizability of our idea to other domains, we also evaluate results on Oxford PET~\cite{PVZJ12} dataset with 24 types of cats and dogs (PET). 
Note that we set the parameters based on experiments on CUB-VW and  used those across all datasets. 

\myparagraph{Experiments on CUB-VWSW dataset.}
We use GFS-EARLY embedding for being the best performing method in our previous evaluation. We compare it with random points, bubbles, bag-of-words and attributes. Results on CUB-VWSW dataset show that gaze performs significantly better than random points that come from the image itself, bubble embeddings extracted from mouse-click locations, and BOW embeddings extracted from Wikipedia articles. On the other hand, expert annotated attributes outperform non-expert annotated gaze data. This is expected since our novice annotators did not compare different vireo, woodpecker, sparrow and warbler species and especially vireos, sparrows and warblers looks very similar to each other, i.e. having similar size, shape and colors. On the other hand, the fact that gaze embeddings perform better than BoW by itself is an interesting result. 
We suspect that allowing the annotators to explore differences between bird species at sub-species level and also super-species level, e.g. our annotators never compared woodpeckers and vireos but only two different woodpecker species, or annotating images using bird expert opinion would improve our results. Additionally, an improved zero-shot learning model that takes into account the hierarchical relationships between classes may lead to better results. We will explore these options in future work. Finally, fine-tuning the parameters of gaze embedding, which we intentionally avoid, may improve results.

\myparagraph{Experiments on PET dataset.}
Here, as attributes and bubbles are not available, we use random points in the images and bag-of-words extracted from Wikipedia articles as baselines. The random chance is $16\%$ as we sample 6 test classes, we repeat our experiments on 10 different zero-shot splits and report the average in the last column of~\autoref{tab:others}. 
We observe that Wikipedia articles of PET classes include more information than random points in the image ($21.0\%$ vs $33.5\%$ with bag of words). Whereas our gaze embeddings obtain $46.6\%$ accuracy that significantly outperforms the results obtained with bag-of-words. As all the images show cat and dog breeds, our annotators are more familiar  with these classes which makes this dataset less challenging than CUB. Note that by fine-tuning raw-gaze processing or gaze embedding parameters such as gaze features, gaze embedding type, etc. on this dataset, these results may potentially get higher. We conclude from PET results that our proposed gaze embeddings indeed capture class discriminative information and they can be generalized to other domains.  

%%%%%%%%%%%%%%%%%%%%%%%%%%%%%%%%%%%%%%%%%%%%%%%%%%%%%%%%%%%%%%%%
\begin{figure*}[t]
   \centering 
    \includegraphics[width=\linewidth]{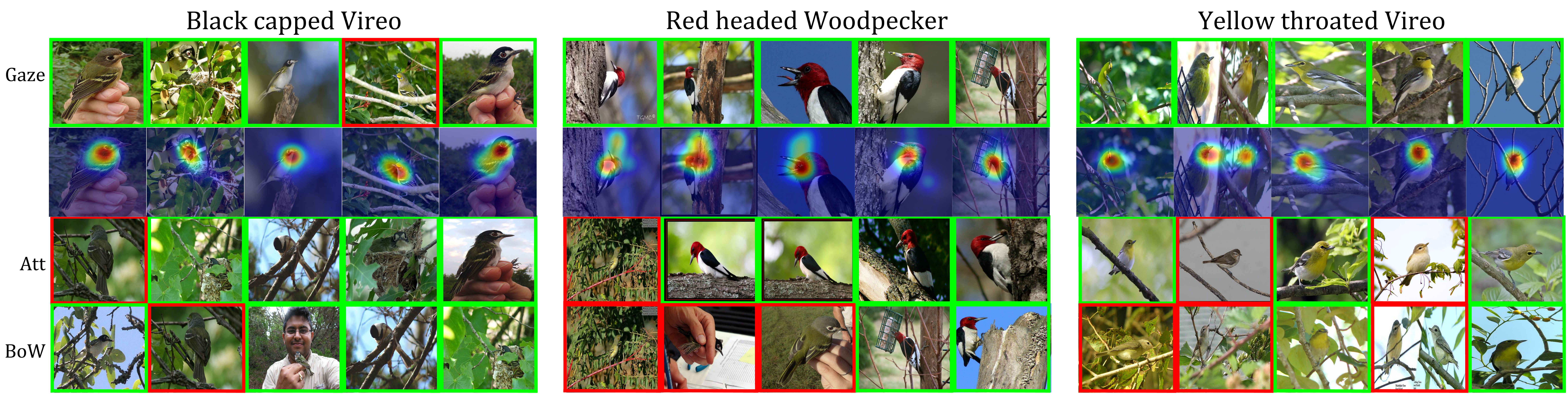}
    \includegraphics[width=\linewidth]{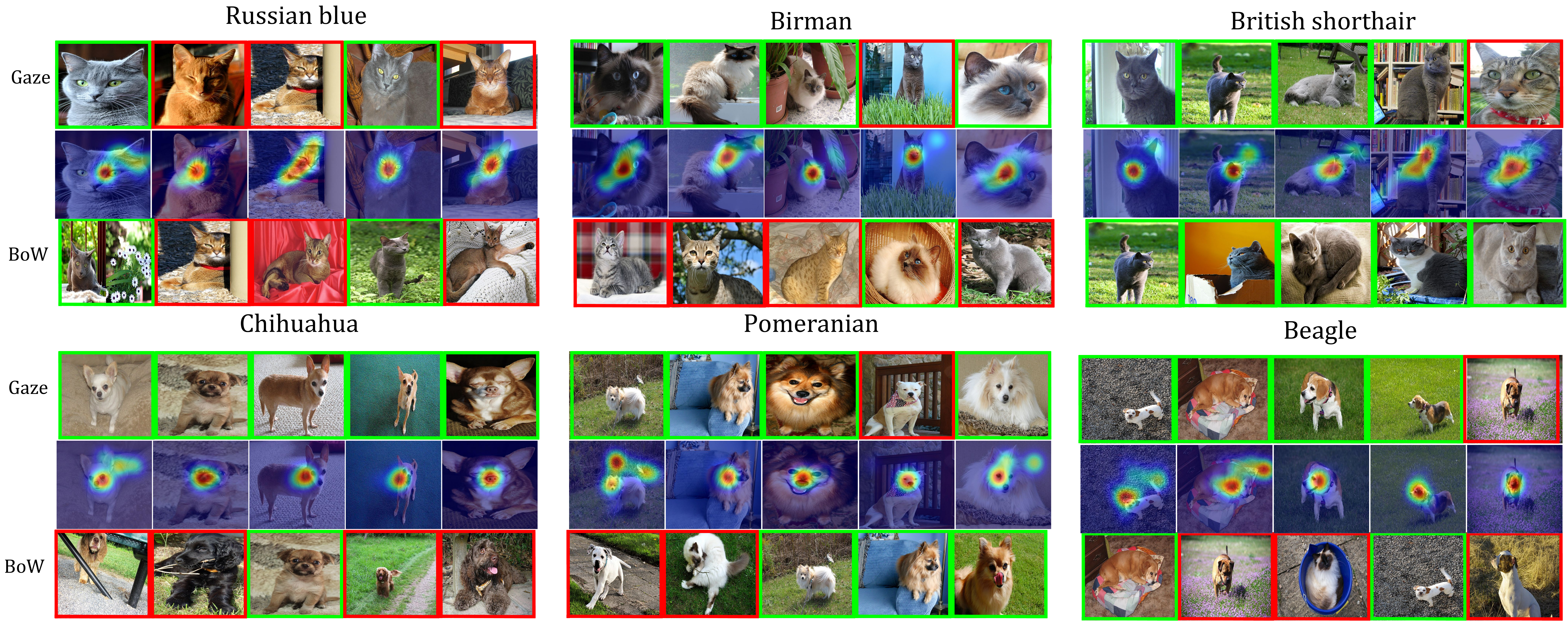}
     \vspace{-2mm}
	\caption{Qualitative Results: Five highest ranked images for unseen classes of birds, cats and  dogs. We compare gaze with attributes (when available) and with bag of word representations and show gaze heat-maps of selected images.}
    \vspace{-2mm}
\label{fig:qualitative}
\end{figure*}  

\subsection{Qualitative Results}
Qualitative results of birds, cats and dogs on  \autoref{fig:qualitative} shows five highest ranked images for three different test classes comparing gaze embeddings with competing methods. We additionally visualize the gaze heatmaps extracted from gaze tracks corresponding to that particular test image. Although we do not use the gaze embeddings for these test classes while training, we include these visualizations as give an intuition of how gaze-tracks look like.

For birds, we compare gaze with both human annotated attributes and bag-of-words. Gaze ranks ``Black capped Vireo'' images correctly in the first three positions whereas attributes and bow makes mistakes. The misclassified ``Black capped Vireo'' is a ``White eyed Vireo'' which also has its distinguishing property on the head region. The misclassified image in expert annotated attributes belongs to ``Blue headed Vireo'' whose embedding is similar to ``Black capped Vireo''. On the other hand, the word 'head' is highly frequent in both Wikipedia articles which makes the BoW embedding for these two classes similar and thus, it leads to a mismatch. For other examples, gaze embedding ranks correct images the highest. These results also illustrate the difficulty of the annotation on fine-grained datasets.

For cats and dogs, we observe that qualitative results follow a similar trend as quantitative results. Qualitatively, gaze performs better than bag-of-words representations. Comparing gaze and bag-of-words results shows that gaze never confuses cats and dogs whereas such confusion occurs for bag-of-words. As a failure case of gaze embeddings, gaze retrieves ``Abyssinian Cat'' images for ``Russian blue'' query as these two cats have a similar form but can be distinguished only with color information, not encoded with gaze.

\section{Conclusion}
\label{sec:conc}

In this work, we proposed to use gaze data as auxiliary information to learn a compatibility between image and label space for zero-shot learning.
In addition to a novel eye tracking data collection that captures humans' natural ability to distinguish between two objects
we proposed three gaze embedding methods that 1) use spatial layout of the gaze points and employ first order statistics, 2) integrate location, duration, sequential ordering and user's concentration features to spatial ordering information, and 3) sequentially sample gaze features.
Through extensive quantitative and qualitative experiments on the CUB-VW dataset we showed that human gaze is indeed class-discriminative and improves over both expert-annotated attributes and mouse-click data (bubbles). 
Our qualitative and quantitative results on the PET dataset showed that gaze can be generalized to other domains. On the other hand, our results on larger fine-grained datasets, e.g. CUB-VWSW might indicate that the results would benefit from alternative data collection paradigms that allow annotators to view super-species as well as sub-species. In future work we will investigate the gaze behavior by focusing on over two fine-grained images. 

\myparagraph{Acknowledgements.} This work was funded, in part, by the Cluster of Excellence on Multimodal Computing and Interaction (MMCI) at Saarland University, Germany. We would like to thank Semih Korkmaz for his helpful insights. 

{\small
\bibliographystyle{ieee}
\bibliography{egbib}
}

\end{document}